\documentclass[10pt,twocolumn,letterpaper]{article}

\usepackage{iccv}
\usepackage{times}
\usepackage{epsfig}
\usepackage{graphicx}
\usepackage{amsmath}
\usepackage{amssymb}

\usepackage{cuted}
\usepackage{subcaption}
\usepackage{url}
\usepackage{color}
\usepackage{enumitem}
\usepackage{comment} 
\usepackage{booktabs}
\usepackage{multirow}
\newsavebox{\largestimage}

\usepackage{mathtools}

\newcommand{\diagentry}[1]{\mathmakebox[1.8em]{#1}}
\newcommand{\xddots}{%
  \raise 4pt \hbox {.}
  \mkern 6mu
  \raise 1pt \hbox {.}
  \mkern 6mu
  \raise -2pt \hbox {.}
}

\makeatletter
\@namedef{ver@everyshi.sty}{}
\makeatother
\usepackage{tikz}
\usetikzlibrary{positioning}

\usepackage[font=small, labelsep=period]{caption} 

\usepackage[pagebackref=true,breaklinks=true,letterpaper=true,colorlinks,bookmarks=false]{hyperref}

\iccvfinalcopy

\def\iccvPaperID{3126}

\newcommand{\change}[1]{#1} 

\begin{document}

\title{EventHands: Real-Time Neural 3D Hand Pose Estimation from an Event Stream\vspace{-5pt}}

\author{ 
Viktor Rudnev$^1$ \hspace{2.5em} 
Vladislav Golyanik$^1$ \hspace{2.5em} 
Jiayi Wang$^1$  \hspace{2.5em} 
Hans-Peter Seidel$^1$  \vspace{5pt} \\ 
Franziska Mueller$^2$ \hspace{2.5em} 
Mohamed Elgharib$^1$ \hspace{2.5em} 
Christian Theobalt$^1$ \vspace{7pt} \\
\hspace{-50pt}
$^1$MPI for Informatics, SIC \hspace{40pt} 
$^2$Google Inc.
}

\twocolumn[{ 
\renewcommand\twocolumn[1][]{#1} 
\maketitle 
\begin{center}
    \includegraphics[width=1.0\textwidth,trim={0 12.5cm 0 0},clip]{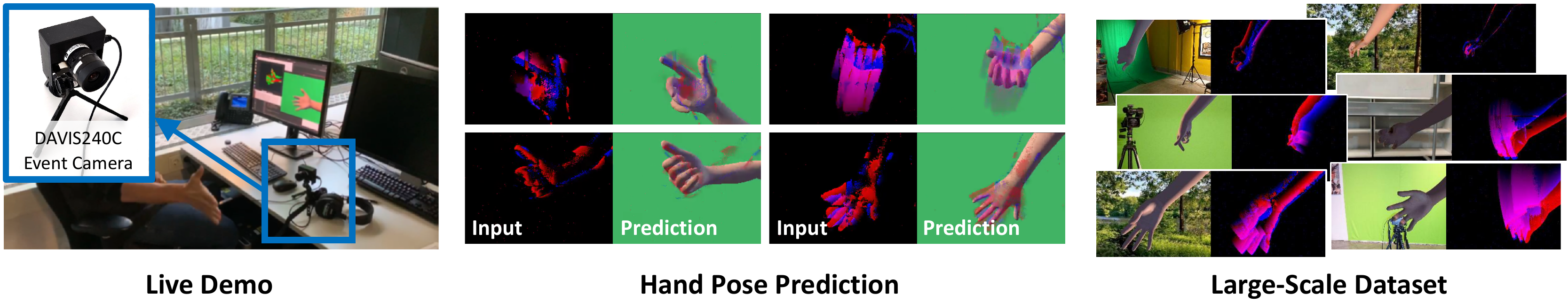} 
    \captionof{figure}{
    Our \emph{EventHands} approach estimates 3D hand poses from asynchronous event streams in real time (no greyscale or RGB images are used at any step of our method). 
    We built a demo system with a DAVIS240C event camera (left) that runs one order of magnitude faster than image-based prior work on 3D hand reconstruction. 
    \emph{EventHands} leverages our new temporal event representation to reconstruct 3D hands in various challenging poses and moving at previously unseen speed (center). 
    Our approach is trained on a large synthetic dataset (right) created by our new highly efficient GPU-based event camera simulator, but generalises well to real data.}
    \label{fig:teaser}
\end{center} 
\vspace{20pt} 
}] 

\maketitle

\begin{abstract} 
   3D hand pose estimation from monocular videos is a long-standing and challenging problem, which is now seeing a strong upturn. 
   In this work, we address it for the first time using a single event camera, \textit{i.e.,} an asynchronous vision sensor reacting on brightness changes. 
   Our \hbox{\emph{EventHands}} approach has characteristics previously not demonstrated with a single RGB or depth camera such as high temporal resolution at low data throughputs and real-time performance at 1000 Hz. 
   Due to the different data modality of event cameras compared to classical cameras,  existing methods cannot be directly applied to and re-trained for event streams. 
   We thus design a new neural approach which accepts a new event stream representation suitable for learning, which is trained on newly-generated synthetic event streams and can generalise to real data. 
   Experiments show that \hbox{\emph{EventHands}}  outperforms recent monocular methods using a colour (or depth) camera in terms of accuracy and its ability to capture hand motions of unprecedented speed. 
   Our method, the event stream simulator and the dataset are publicly available (see  \url{https://4dqv.mpi-inf.mpg.de/EventHands/}). 
\end{abstract} 

\section{Introduction}\label{sec:introduction} 
    Event cameras are vision sensors which respond to events, \textit{i.e.,} local  changes in the incoming brightness signals. 
    In contrast to conventional RGB cameras which record images at a pre-defined frequency (\textit{e.g.,} $30{-}60$ fps), event cameras operate asynchronously, 
    which enables high clock speeds and temporal resolution of up to $1\mu s$ \cite{Lichtsteiner2008}. 
    Due to their unique properties and high-dynamic range, event cameras have already found applications in 
    low-level vision \cite{Bardow2016, Pan2019, Rebecq2020, Zhang2020}, low-latency robotics \cite{Sugimoto2020, Falanga2020}, visual simultaneous localisation and mapping (SLAM)  \cite{Weikersdorfer2013, Kim2016}, feature and object tracking \cite{AlzugarayChli2018, Mitrokhin2018}, gesture recognition \cite{Amir2017, Wang2019}, experimental physics (particle tracking and velocimetry) \cite{Borer2017, Wang2020} and astronomy \cite{Chin2019, Zolnowski2019}, among other fields. 
    
    In this work, we are interested in 3D hand pose regression using a single event camera. 
    The high dynamic range, lower latency, and lower throughput of event data are better suited than conventional images for tracking hands, which are often in rapid motion.
    However, due to the drastically different and less regular data from event cameras, 
    existing RGB- or depth-based methods~\cite{Boukhayma_2019_CVPR, Zhang_2019_ICCV, HandVoxNet2020, GANeratedHands_CVPR2018} cannot be directly applied to event streams. 
    A na{\"i}ve way would be to reconstruct greyscale images from an event stream at arbitrary 
    temporal resolutions first, and then run the same monocular methods \cite{Baek_2019_CVPR, Boukhayma_2019_CVPR, Zhang_2019_ICCV, Zhou2020, Qian2020} on those. 
    This policy would, unfortunately, nullify most advantages of event cameras such as low data bandwidth, abstraction from a large variety in textures and illumination conditions and supposedly better generalisation ability. 
    It is also not clear how existing methods would perform on low-resolution greyscale images reconstructed from event streams, as the latter 
    often contain artefacts, and cannot reproduce the exact occurred brightness, due to non-deterministic event thresholds and noise \cite{Bardow2016, Reinbacher2016, Scheerlinck18}. 
    The primary research question is thus how
    hands can be reconstructed and tracked in 3D directly from event streams. 
    
    We pursue in this paper a learning-based approach and propose the first---to the best of our  knowledge---method for 3D reconstruction of a human hand from a single event stream (see Fig.~\ref{fig:teaser}). 
    Our neural \emph{EventHands} approach learns to regress 3D hand poses, represented as global rotation and translation as well as pose parameters of a parametric hand model~\cite{Romero2017}, from \emph{locally-normalised event surfaces} (LNES), which is a new way 
    of accumulating events 
    over temporal windows for learning. 
    We generate training data for our neural network using a new high-throughput event stream simulator relying on a parametric hand model \cite{Romero2017, Qian2020}. 
    The training data includes variations in hand shape and texture, lighting, and scene background, and accurately mimics the characteristics of a real event camera. 
    Hence, \emph{EventHands} generalises well to real data despite being trained with synthetic data only. 
    Next, \emph{EventHands} runs at $1$ KHz, which is significantly faster than any image-based prior works. 
    In summary, our contributions are: 
    \begin{itemize}\itemsep0em
        \item \emph{EventHands}, \textit{i.e.,} the first approach for 3D hand pose estimation, including rotation and translation in 3D, from a single event stream, running at 1 KHz. 
        \item A new high-throughput event stream simulator 
        supporting a parametric 3D hand model for diverse poses, shapes, and textures, multiple light sources, adjustable event stream properties  
        (\textit{e.g.,} event threshold distributions, noise patterns)
        and further scene augmentation. 
        \item A live real-time demonstrator of our method running orders of magnitude faster than previous image-based work on a workstation with a single GPU. 
        See our supplementary video for recordings thereof. 
    \end{itemize} 
    We evaluate the proposed approach on a wide variety of motions with real and synthetic data, and provide numerical evidence for our design choices as well as comparisons to prior work. 
    We show that \emph{EventHands} yields accurate estimates even when existing RGB- and depth-based techniques fail due to fast motion. 

\section{Related Work}\label{sec:related_work} 
    We next review related works on 3D hand reconstruction and event-based vision. 
    Our \emph{EventHands} is the first approach for 3D hand pose estimation operating on event streams and has multiple advantages compared to existing RGB- or depth-based methods highlighted in the following. 
    
    \noindent\textbf{3D Hand Reconstruction Methods.} 
    The vast majority of existing works for 3D hand reconstruction from depth~\cite{tompson_tog2014,Oberweger2015,wan2017crossing,Ge2018,Moon2018,LiLee2019,Fang2020} and monocular RGB~\cite{Zimmermann2017,cai2018weakly,Spurr2018,GANeratedHands_CVPR2018,Yang2019,Spurr2020} regresses sparse hand joints. 
    Several recent works also address dense 3D reconstructions of hands \cite{Baek_2019_CVPR, Boukhayma_2019_CVPR, Zhang_2019_ICCV, Zhou2020, Malik2018, HandVoxNet2020, Wan2020, Shen2020, Qian2020}, some of which rely on a parametric 3D hand model such as MANO~\cite{Romero2017} for pose and shape or HTML~\cite{Qian2020} for textures.
    Hampali~\textit{et al.}~\cite{hampali2020honnotate} introduces a new benchmark for hand-object interaction methods. 
    The dataset is then leveraged for 3D hand pose estimation from RGB images by fitting the MANO model to the predicted 2D hand joints. 
    Taylor \textit{et al.}~\cite{Taylor2017} proposed an approach for hand tracking from a new custom-built depth sensor.
    Their custom depth camera supports $180$ fps which is significantly faster than commodity depth cameras (30--60fps) but still far from the temporal resolution of an event camera. 
    
    All the aforementioned methods cannot be directly applied to event streams. 
    Even though intensity images and videos can be reconstructed from event streams \cite{Reinbacher2016, Scheerlinck18}, the obtained greyscale images considerably differ from the data used by existing hand reconstruction techniques and may exhibit
    domain-specific artefacts.
    Bridging this domain gap is not straightforward.
    On the other hand, direct operation on event streams has the advantage of low data bandwidth and abstraction from appearance variation occurring in RGB images. 
    
    \noindent\textbf{Event-Based Vision Techniques.} 
    Since dynamic vision sensors or event cameras became available
    , they have been predominantly used for low-level and mid-level problems such as greyscale image restoration from events \cite{Pan2019, Zhang2020}, optical flow \cite{Bardow2016, Pan2020}, or feature detection and tracking \cite{Vasco2016, AlzugarayChli2018, Mitrokhin2018}. 
    In the context of our work, noteworthy are SLAM methods \change{\cite{Weikersdorfer2013, Weikersdorfer2014, Yuan2016, Kim2016, Zhou2021}} which rely on sparse rigid 3D reconstruction as an auxiliary task to localise moving robots, event-driven stereo matching \cite{Schraml2015, Rebecq2018, Zhou2018}, and 2D gesture recognition \cite{Amir2017, Wang2019}.
    To apply learning-based method on event streams, suitable representations for the input have been investigated, for example, event frames~\cite{Rebecq2017}, event count images~\cite{Maqueda2018,  Zhu2018}, surfaces of active events (SAE)~\cite{Benosman2014}, time-surfaces~\cite{Lagorce2017}, \change{hierarchy of time surfaces~\cite{Lagorce2017}}, averaged time surfaces~\cite{Sironi2018}, \change{sorted time surfaces~\cite{Alzugaray2018},} and differentiable event spike tensors~\cite{Gehrig_2019_ICCV}, among others. 
    Our LNES representation relates to SAE and differs from it by representing time stamps in window-normalised time units. 
    \change{For a more detailed discussion of event representations, please refer to the survey by Gallego \textit{et al.}~\cite{Gallego2020}.}
    
    A related work to ours which tracks a human in 3D from a hybrid input of events and greyscale images is EventCap~\cite{EventCap2020CVPR}. 
    It relies on event correspondences between greyscale anchor frames and assumes a known rigged and skinned human body template. 
    Compared to human bodies, hands exhibit much more self-occlusions, which makes it difficult to obtain event trajectories or perform image-based model fitting. 
    \change{Nehvi \textit{et al.}~\cite{Nehvi2021} propose an unsupervised learning approach for deformable object tracking in 3D, which correlates observed and simulated event streams. 
    However, their method requires an accurate initialisation, supports only simple hand motions, and operates far from real time.} 
    Instead, we train a neural network to regress challenging 3D hand poses directly from an event representation suitable for learning (LNES), enabling live applications running five orders of magnitude faster compared to \cite{EventCap2020CVPR, Nehvi2021}. 
    \change{
    Although there exist general-purpose event camera  simulators~\cite{kaiser2016towards, koenig2004design,  rebecq2018esim}, we develop a new hands-specific simulator for generating training data. 
    This has the advantage that the parametric hand model is tightly integrated into it, enabling on-the-fly sampling of realistic textures, poses and shapes. %
    Moreover, it is tailored for a high data generation speed 
    with seamless GPU support. 
    } 
    
    All in all, our \emph{EventHands} approach further advances the underexplored area of 3D reconstruction and tracking of non-rigid objects from events.

\section{Event Camera Model}\label{sec:overview}

    While event cameras obey the pinhole camera model of geometric projections from the 3D space to the 2D image plane, each pixel of an event camera independently and asynchronously reacts to differences in the observed logarithmic brightness $\mathcal{L}(u, t)$. 
    An event $e_i = (u_i, t_i, p_i)$ is a $3$-tuple with the pixel identifier $u_i$, triggering time $t_i$ and the binary polarity flag $p_i \in \{-1, 1\}$ signalising whether the logarithmic brightness has increased or decreased by an absolute threshold $|C|$, \textit{i.e.,} an event is triggered at time $t_i$ as soon as one of the following two conditions is satisfied: 
    \begin{equation} 
    \begin{cases} 
    \mathcal{L}(u_i, t_i) - \mathcal{L}(u_i, t_p) =  \;\;\,C & (p =  \;\;\,1) \\ 
    \mathcal{L}(u_i, t_i) - \mathcal{L}(u_i, t_p) = -C & (p = -1) 
    \label{eq:event_gen}
    \end{cases}, 
    \end{equation} 
    where $t_p$ is the previous triggering time of an event at $u_i$.
    The event camera we use in our experiments (DAVIS240C) provides the microsecond precision for $t_i$. 
    Due to the hardware reasons, $C$ is not a fixed threshold but rather follows an unknown distribution of the thresholds $\chi$. 
    In event camera modelling, it is, however, convenient and sufficient to assume $C$ to be equal to the expected value of $\chi$. 
    Moreover, a capacitor attached to each sensor pixel can suddenly overfill,  which leads to a spurious noise event registration.

\section{Event Stream Simulator and the Dataset}\label{sec:simulator} 

Due to the lack of event stream dataset for hand pose estimation, and the difficulty of obtaining accurate 3D ground-truth annotations on real data, we build a highly efficient event stream simulator to generate a large-scale synthetic event stream dataset with annotations.
In total, we generated $100$ hours of simulated event data, which provides $3.6 \cdot 10^8$ discrete time steps with ground-truth annotations for training. 
This places our data among the most extensive event stream datasets available for research purposes so far. 
We plan to release both the dataset and the simulator.

\subsection{Scene Modelling}
    
\noindent\textbf{Hand and Arm Model.} 
Our simulator models both arm and hand geometry, as modelling the hand alone would generate spurious events generated from the seam at the wrist.
Hence we use SMPL+H ~\cite{Romero2017}, \textit{i.e.,} a model that combines the hand model MANO ~\cite{Romero2017} and the body model SMPL ~\cite{SMPL:2015}. 
To capture events generated by the hand appearance, we use the texture model of Qian \textit{et al.}~\cite{Qian2020} for the hand.
The arm texture is obtained by extending the average hand boundary colour to the rest of the SMPL+H mesh. 

\noindent\textbf{Model Animation.} 
To simulate hand articulation, we sample individual model poses using the provided MANO PCA-based parameter spaces to obtain a natural distribution of hand poses. 
Additional random offsets to the translation and pose parameters of SMPL+H are added to account for rigid body transformations of the hand and to increase variations in  arm events. 
To generate plausible motion, we select a new random pose every single simulated second and smoothly interpolate between those poses using a quadratic Bezier curve. The curve's middle control point is also randomly selected every single simulated second. This ensures that every second, there would be a sharp change in the motion direction 
(\textit{e.g.,} as in hand waving motions).

\noindent\textbf{Lighting Model.} 
We use a Lambertian lighting model with two lights.
Suppose that $n \in \mathbb{R}^{3}$ is the normal vector to the object surface,
$l_1, l_2 \in \mathbb{R}^{3}$ are light directions and 
$c_1, c_2, c_{\textrm{ambient}} \in \mathbb{R}^{3}$ are light colours. 
Then,
\begin{equation} 
\begin{aligned}
    &\text{light} =  \langle n, l_1 \rangle c_1 + \langle n, l_2 \rangle c_2 + c_{\textrm{ambient}},  \\
    &\text{linear colour} = \textrm{light} \odot \textrm{albedo}, 
\end{aligned} 
\end{equation}
where $``\odot"$ denotes element-wise product. 

\noindent\textbf{Image Formation.}  The scene model and the light model are used to form an RGB image $F_i \in [0,255]^{W\times H\times3}$ at time $t_i$. 
We convert $F_i$ to a log-brightness image $\mathcal{L}(t_i) \in \mathbb{R}^{W\times H}$ using the estimate 
\begin{equation}
    \mathcal{L}(t_i) = \log (0.2F_i^r+0.7F_i^g+0.1F_i^b + \epsilon), 
\end{equation} 
where $\epsilon = 1.0$ is added for numerical stability, and  $F_i^r$, $F_i^q$ and $F_i^b$ are red, green and blue image channels,  respectively. %

\subsection{Event Camera  Simulation}\label{ssec:event_camera_simulator} 

\begin{figure*}[ht]
    \centering 
    \includegraphics[width=1.0\textwidth,trim={0 12cm 0 0},clip]{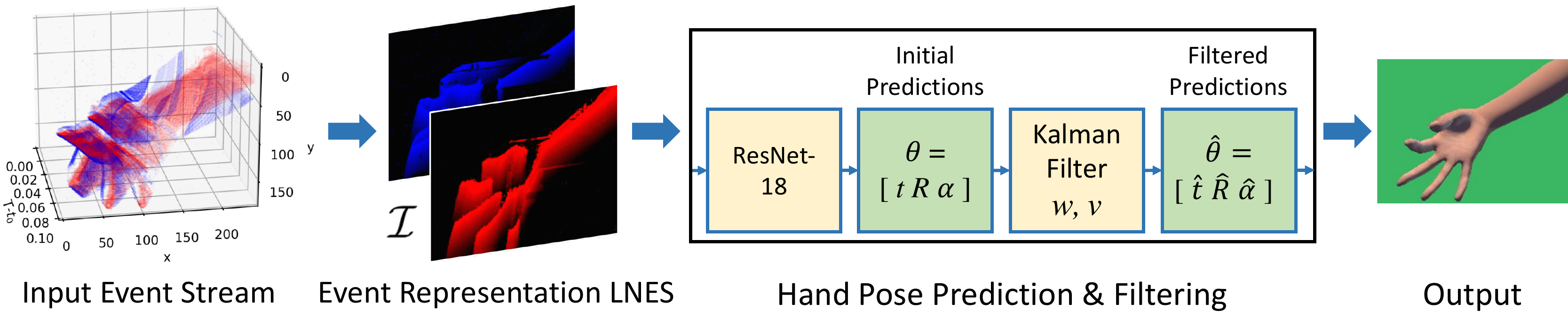}
    \caption{
    \textbf{\emph{EventHands} method overview.} 
    Our approach converts temporal windows of events to the LNES representation with two channels for positive and negative events. 
    The Hand Pose Prediction and Filtering stage uses a neural network (ResNet-18) and a Kalman filter to estimate hand pose as well as the hand translation and rotation. 
    The output shows the rendered hand shape proxy with the estimated parameters. 
    The neural network is trained on our new large-scale dataset for hand pose estimation from event streams. 
    }
    \label{fig:method_overview} 
\end{figure*} 

\noindent\textbf{Event Stream Generation.} 
To simulate events at a time $t_i$, at each pixel location $u_i$, we extract log-brightness $\mathcal{L}(u_i, t_p)$. 
We additionally maintain a memory frame $M \in \mathbb{R}^{W\times H}$, where $M(u_i) \approx \mathcal{L}(u_i, t_p)$ is the absolute log-brightness of the last generated event at $u_i$ at time $t_p$. 
An event tuple $(u_i, t_i, p_i)$ is simulated using the following steps: 
\renewcommand{\labelenumii}{\theenumii}
\renewcommand{\theenumii}{\theenumi.\arabic{enumii}.}
\begin{enumerate}[topsep = 3pt, itemsep = 1pt, partopsep = 3pt, parsep = 3pt] 
  \item Noise events: Emit an event tuple with positive or negative polarity with probability $p_\mathrm{positive}$ and  $p_\mathrm{negative}$, respectively. 
  \item Calculate the logarithmic brightness difference $\Delta =  \mathcal{L}(u_i, t_i) -M(u_i)$, and 
   \begin{enumerate}
   \item If $\Delta \geq C$, emit $\lfloor \Delta/C \rfloor$ positive events. Update the memory frame $M(u_i) = M(u_i)+\lfloor \Delta/C \rfloor C$. 
  \item If $\Delta \leq -C$, emit $\lfloor -\Delta/C \rfloor$ negative events. Update $M(u_i) = M(u_i)-\lfloor -\Delta/C \rfloor C$. %
   \end{enumerate}  
\end{enumerate}
The threshold $C$ and the noise event rates $p_\mathrm{positive}$, $p_\mathrm{negative}$ are calibrated to match our DAVIS240C event camera. 
See supplementary material for the details. 

\noindent\textbf{Data Augmentation.}
Augmentation is critical for successful synthetic-to-real domain transfers~\cite{openairandom}. 
Thus, we re-randomise most of the aspects of the simulation every 50 simulated seconds. 
Those are the hand and body shapes, body position, hand texture, light directions and intensities, background image and its cropped region, as well as $C$. 
Please refer to the supplementary document for the randomisation ranges for each variable. 

\noindent\textbf{Simulation Runtime.} 
The simulation is capable of rendering and extract events from around $2000$ log-brightness images per second. Using temporal resolution of 1000 fps, this allows us to generate ${\approx}100$ hours of simulated event data in two days on a single NVIDIA GTX 1070 GPU. 
Fig.~\ref{fig:teaser} (right) shows sample data synthesised using our simulator. 

\section{Proposed Approach}\label{sec:approach} 

In the previous section, we introduced the \emph{Event\-Hands} dataset generated by our new event stream simulator. 
We now describe our neural approach for 3D hand pose prediction from event streams for which an overview is shown in Fig.~\ref{fig:method_overview}. 
We first describe our event stream representation for learning  (Sec.~\ref{ssec:representation_for_learning}). 
Next, we elaborate on our method which consists of two stages, \textit{i.e.,} \textit{hand pose prediction}   (Sec.~\ref{ssec:hand_pose_prediction}) and \textit{temporal filtering} (Sec.~\ref{ssec:temporal_filtering}).

\subsection{Our Representation of Events for  Learning}\label{ssec:representation_for_learning} 
The original event stream output of an event camera is an asynchronous and 1D. 
At the same time, most recent advances in visual machine learning have explored models that work on spatial 2D images, 3D voxel grids or graphs. 
The straightforward way to convert the 1D event stream to a 2D representation is to accumulate and collapse all events in a time interval, which leads to the loss of temporal resolution within the interval~\cite{Maqueda2018}. 
Hence, we propose a 2D %
representation called Locally-Normalised Event Surfaces (LNES), which encodes all events within a fixed time window as an image $\mathcal{I} \in \mathbb{R}^{W\times H\times 2}$ (see Fig.~\ref{fig:method_overview}, left). 
Using separate channels for positive and negative events preserves the polarities and reduces the number of overridden events. 
In contrast to existing representations, (\textit{e.g.,} \cite{Benosman2014}), LNES operates with window-normalised time stamps. 

Consider the $k$-th time window in an event stream of size $L$. 
We can create the LNES representation of this window, $\mathcal{I}_k$, by first initializing it with zeros, and  
collect events $\mathcal{E} = \{(t_i, x_i, y_i, p_i)\}_{i=1}^{N_k}$ which have timestamps $t_i$ within this window. 
$\mathcal{I}_k$ is updated by iterating through $\mathcal{E}$ from the oldest to the newest event and assigning 
\begin{equation} 
    \mathcal{I}(x_i, y_i, p_i) = \frac{t_i-t_0}{L}.  
\end{equation} 
Thus, $\mathcal{I}(x_i, y_i, p_i) \ne 0$ is a window-normalised timestamp of the event which preserves the relative temporal correlation of the events within the window. 
Note that due to the iteration order $\mathcal{I}(x_i, y_i, p_i)$ can be overridden when a new event with the same polarity is occurring at the same pixel.
For our experiments, we used a fixed time length window of 100ms with a 99ms overlap between consecutive windows. 
Hence, our representation has an effective temporal resolution of $1$ $ms$ to match the inference speed of our network.

The proposed event stream representation allows for several augmentations.
For example, different contrasts between skin tone and background colour can be simulated by switching the polarity of some events. 
This can be easily performed in LNES by swapping the content of the two channels at a subset of pixels.
We also augment the speed of the motion during training by changing the window length without having to re-generate a  dataset with new settings. 

Note that in contrast to na\"{i}ve event accumulation in a time window  \cite{Rebecq2017, Maqueda2018}, where the temporal ordering of the events within the window is lost, LNES preserves temporal information of the events which leads to a more expressive input for learning.
In addition, this enables our method to run with large window sizes without losing temporal resolution and hence prediction quality.
In Sec.~\ref{sec:ablation}, we provide experimental evidence for the merits of our representation.

\subsection{Hand Pose Prediction} 
\label{ssec:hand_pose_prediction} 
We represent the hand pose with $\theta = [ t, R,\alpha ]  \in \mathbb{R}^{12}$, where $\alpha \in \mathbb{R}^{6}$ are the coefficients of the MANO PCA pose space, and $t, R \in \mathbb{R}^{3}$ encode the rigid translation in meter and the rotation in axis-angle formulation, respectively. 
Note that we assume constant lighting and static background relative to the event camera, \textit{i.e.,} all events are due to the hand and arm, up to noise. 

We train a ResNet-18~\cite{resnet} on our event input representation $\mathcal{I}$ to regress the pose representation $\theta$. 
This architecture allows us to predict ${\approx}750{-}1550$ poses per second depending on the GPU (GTX 1070 \textit{vs} RTX 2080 Ti) to fully take advantage of the millisecond temporal resolution LNES. 
Please refer to the supplementary document for a more detailed discussion of network architecture choice. 
During training, we minimise the following loss function $\mathcal{L}$: 
\begin{equation} 
    \mathcal{L} = \mathcal{L}_{\alpha} + \lambda_t \mathcal{L}_{t} + \lambda_R  \mathcal{L}_{R}, 
\end{equation} 
with the MANO loss $\mathcal{L}_{\alpha}$, translation loss $\mathcal{L}_{t}$, rotation loss $\mathcal{L}_{R}$ (all $\ell_2$ losses) along with the weights $\lambda_t = 500.0$ and $\lambda_R = \frac{1}{3}$. 
These weights are chosen empirically for normalisation to account for parameters of  different magnitudes. 
We use the most significant MANO components corresponding to the six largest eigenvalues during training and inference. 
We use $45$ hours of synthetically-generated event streams to train our neural network. 

\subsection{Temporal Filtering}
\label{ssec:temporal_filtering}

Although our novel input representation explicitly models relative temporal information of the events within the event window, and we use overlapping event windows for pose prediction, sequences of raw network predictions still exhibit temporal jitter due to missing longer-term correlation across event windows.
This is especially relevant on real test data where jitter is more severe due to the domain gap (see Sec.~\ref{sec:ablation}).
Hence, we apply additional temporal filtering by using a constant-velocity Kalman filter \cite{Kalman1960} on the raw network outputs. 
We set the process noise $W = \omega(0.1)$ and the observation noise $v = 5.0$ for low-speed movements 
and $W =\omega(3.0)$ and $v = 1.0$ for high-speed movements, 
with $\omega(\cdot)$ being discrete white noise covariance matrix operator  \cite{labbe2014kalman}. 
See our supplement for the exact form of $\omega(\cdot)$. 

\section{Results}\label{sec:experiments} 

\begin{table}
    \centering 
    {
\footnotesize{
\begin{tabular}[b]{llccc}
\toprule
    & & \multicolumn{2}{c}{synthetic} & real \\
    \cmidrule(lr){3-4}
    \cmidrule(lr){5-5}
     & & \textbf{2D-AUCp} & \textbf{3D-AUC} & \textbf{2D-AUCp} \\
     \midrule
    & no filtering & \textbf{0.89} & \textbf{\textit{0.85}} & \textbf{\textit{0.75}} \\ 
    & no aug. & \textbf{\textit{0.88}} & \textbf{0.86} & 0.70 \\ \midrule
    \multirow{2}{*}{EOI} & 33ms & 0.86 & \textbf{\textit{0.85}} & 0.70 \\
    & 100ms & 0.78 & 0.80 & 0.56 \\ \midrule
    \multirow{2}{*}{ECI-S} & 33ms & \change{0.83} & \change{0.81} & 0.66\\
    & 100ms & \change{0.69} & \change{0.76} & 0.56\\ \midrule
    \multirow{2}{*}{ECI} & 33ms & \change{0.86} & \change{0.83} & 0.69\\
    & 100ms & \change{0.76} & \change{0.79} & 0.52\\ \midrule
    \multirow{3}{*}{LNES} & 33ms & \textbf{\textit{0.88}} & \textbf{\textit{0.85}} & 0.72 \\
    & 300ms & 0.87 & 0.84 & 0.72 \\
    & \textbf{proposed} & \textbf{\textit{0.88}} & \textbf{\textit{0.85}} & \textbf{0.77} \\
    \bottomrule
\end{tabular}
}
}
    \caption{
        Ablation study on synthetic and real test data. 
        We report the 2D-AUCp and the 3D-AUC (higher values are better, \change{bold/bold italic font denotes best/second-best numbers)}. 
    }
     \label{fig:ablation_table}
\end{table}

We perform experiments on multiple sequences and demonstrate the ability of our approach in capturing a wide variety of motions, including translation, rotation and articulations. 
\emph{EventHands} is able to accurately reconstruct hands that are moving at speeds previously unseen in the literature. 
We first introduce our evaluation metrics and test data (Sec.~\ref{sec:metrics}).
Then we present evaluations of our design choices (Sec.~\ref{sec:ablation}), compare against related techniques~\cite{Zhou2020,GANeratedHands_CVPR2018,Boukhayma_2019_CVPR,Moon2018} 
(Sec.~\ref{sec:comparisons}) and provide additional results of our method (Sec.~\ref{sec:additional_results}). 
For more visual results and comparisons, please refer to the supplemental video. 

We visualise results on a mean hand shape~\cite{Romero2017} with a mean texture~\cite{Qian2020}. 
Note that our work focuses on predicting the motion of only the hand, and not the arm. 
For visualisation purposes, we render the arm using the predicted  parameters to ensure it attaches to the predicted hand. 
Nevertheless, this can produce arm movement different from the ground truth. 
We kindly ask readers to ignore the predicted motion of the arm as it is outside the scope of our work.    

\subsection{Metrics and Test Data}
\label{sec:metrics}
\noindent\textbf{Synthetic Data.} 
For the synthetic test set, we simulate a total of $1240$ seconds of hand motions with $2.64 \cdot 10^8$ events. 
Ground-truth annotations on all $21$ keypoints are available at 1ms intervals.

\noindent\textbf{Real Data.}
We recorded four real event sequences totalling $12\,600$ milliseconds and 
$5.93 \cdot 10^6$ events using the DAVIS240C event camera and a single synchronised high-speed RGB camera Sony RX0.
Considering each LNES event window sampled at 1ms temporal resolution as a frame, the sequences are annotated evenly at $30$ frames per second for a total of $357$ frames.

To obtain 2D annotations, we first use OpenPose \cite{OpenPose,simon2017hand} on the $500$~fps high-speed RGB footage to generate initial reference keypoints.
The keypoints for fingertips, middle MCP, and wrist are then manually inspected and corrected by multiple annotators to obtain ground-truth annotations.
Similarly, hand keypoints are manually annotated on event images from the real event streams and in a second step inspected and corrected by a second set of annotators. 
In total, we obtained $2499$ keypoints over $357$ frames.
Please refer to the supplemental video for visualizations of the annotation quality.

\begin{figure}
    \includegraphics[width=\linewidth]{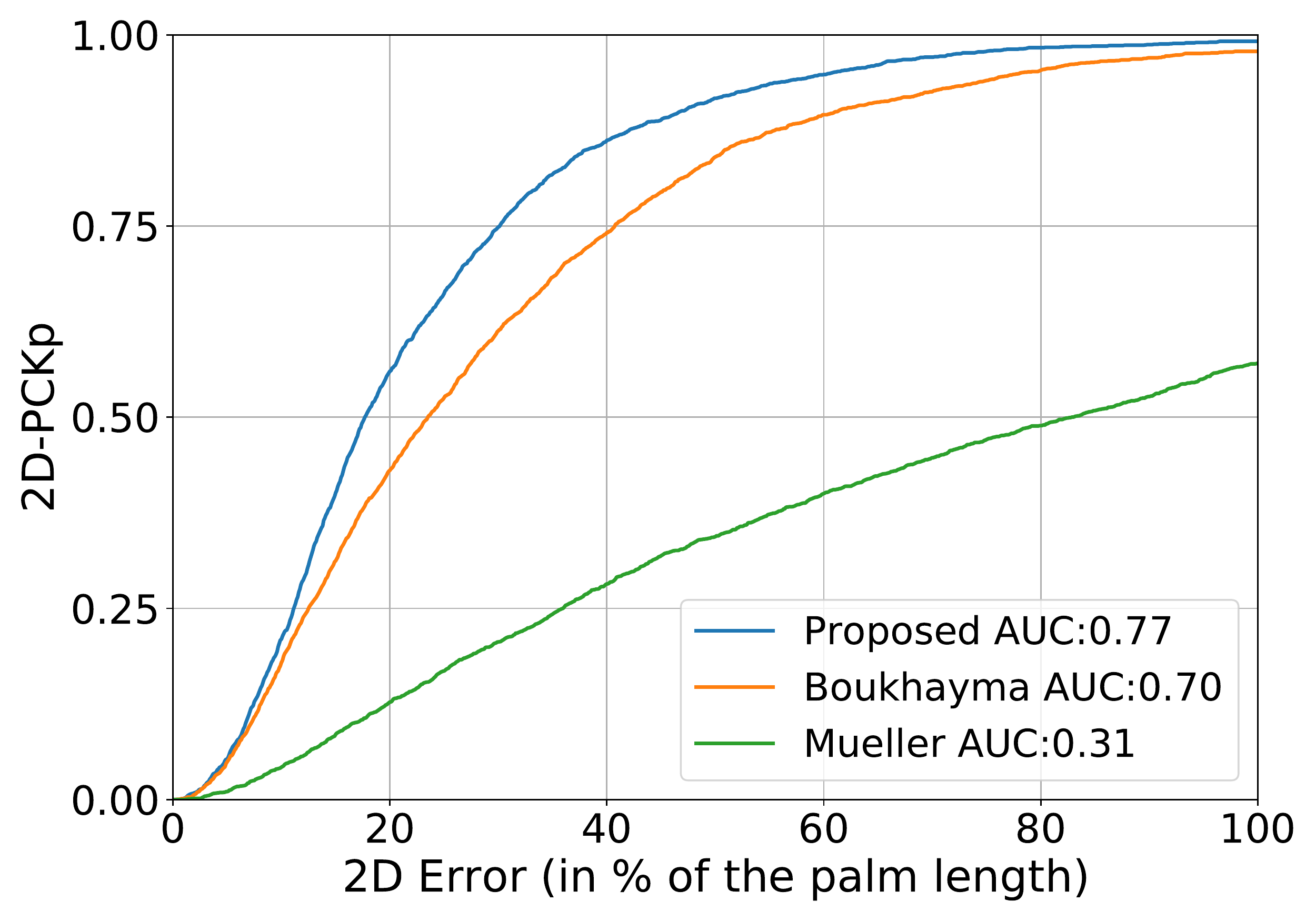}
    \caption{
        Quantitative results of the RGB-based hand pose estimation methods by Mueller \textit{et al.}~\cite{GANeratedHands_CVPR2018} and Boukhayma \textit{et al.}~\cite{Boukhayma_2019_CVPR}. 
    }
    \label{fig:comparison_real}
\end{figure}

\noindent\textbf{Evaluation Metrics.} 
When 3D keypoints are available, we evaluate the root-aligned percentage of correct 3D keypoints (3D-PCK)~\cite{GANeratedHands_CVPR2018} and the area under the PCK curve (3D-AUC) with thresholds ranging from $0$ to $100$mm. 

For real data, we cannot calculate the 3D-PCK since we do not have access to ground-truth 3D annotations and obtaining them manually is challenging. 
Instead, we report the 2D-PCK and the corresponding area under the curve (2D-AUC).
To make the 2D-PCK comparable across different data modalities when comparing to existing RGB methods, we use the wrist and middle finger MCP annotations to calculate the average palm length in pixels for each sequence and normalise the 2D errors by it. 
Analogously to the 2D body pose estimation literature~\cite{andriluka20142d}, we refer to the palm-normalised 2D-PCK as \emph{2D-PCKp} and the corresponding AUC as \emph{2D-AUCp}.
Here, we use a threshold ranging from $0$ to $100$\% of the relative palm length.

\begin{figure*}[ht]
    \centering 
    \input{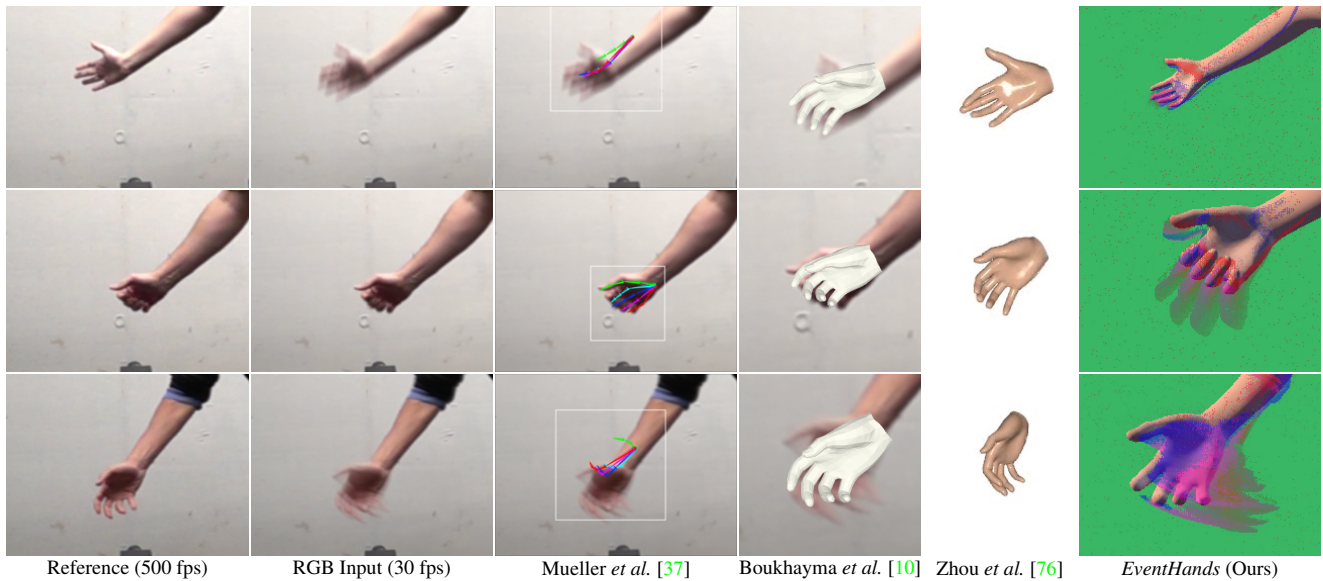}
    \vspace{-0.3cm}
    \caption{
    Comparison against state-of-the-art RGB hand pose estimation techniques. 
    We show the high-frame-rate footage (first column) only for reference, while the RGB techniques process a version downsampled to 30fps (second column). 
    Mueller~\textit{et al.}~\cite{GANeratedHands_CVPR2018} estimates wrong bounding boxes and hence produces erroneous network predictions which are propagated to their final IK skeleton fit (shown here). 
    Boukhayma \textit{et al.}~\cite{Boukhayma_2019_CVPR} estimates wrong rigid rotation on blurry input and often resorts to approximately the MANO mean pose (first and last row). 
    Zhou~\textit{et al.}~\cite{Zhou2020} do not estimate any hand translation and hence cannot handle translational motion (first rows).
    Furthermore, their method struggles with fast blurry motion (last row). %
    Our approach produces accurate 3D hand poses including global translation and rotation, also for challenging articulations like fists (second row) and clearly outperforms state of the art, especially on fast motions. 
    } 
    \label{fig:compRGB} 
\end{figure*} 

\subsection{Ablation Study}
\label{sec:ablation}

We quantitatively evaluate the different design choices of our technique on synthetic and 
real test data (Table~\ref{fig:ablation_table})
Note that the synthetic test data is an easier setting for our method since it was trained on synthetic event streams. 
Hence, different versions achieve similar results. 
The tests on real sequences which shows the benefits of our design choices for generalisation purposes.

\noindent\textbf{Influence of Data Augmentation.}
As discussed in Sec.~\ref{ssec:event_camera_simulator}, we use several data augmentation schemes to diversify our event stream data used for training.
Augmentation does not help on synthetic data since there is no domain gap to be bridged.
On real data, using data augmentation significantly improves the quality of the predictions.

\noindent\textbf{Influence of Temporal Filtering.} 
We use a Kalman filter to improve the longer-term temporal smoothness of our predictions (see Sec.~\ref{ssec:temporal_filtering}).
On synthetic data, both versions perform similar (Table~\ref{fig:ablation_table}).
On real data, however, where temporal jitter is larger due to the domain gap, the proposed filtering improves the results.
Since temporal smoothness is best examined in videos, we refer to our supplemental video for visual results of this ablation study. 

\noindent\textbf{Influence of Input Event Representation.}
We compare to three different event representations as baselines: event occurence images (EOI), single-channel event count images (ECI-S)~\cite{Rebecq2017}, and two-channel event count images (ECI)~\cite{Maqueda2018}.
EOI and ECI consist of two channels, one for each polarity.
EOI contain binary event occurence flags for each pixel whereas ECI contain the accumulated number of all events occured for each pixel in the time window.
ECI-S is a simpler version of ECI where all events are accumulated in a single channel irrespective of their polarity.
For more details about the baselines, please refer to the supplemental document.
In contrast to our LNES, these other event representations do not consider the temporal information of the events.
\change{The best window size is task-specific. For an event representation it is hence advantageous to support a wide range of window sizes. Our evaluation shows that LNES captures meaningful and precise information without degrading for longer windows (where more events are condensed).}
\emph{EventHands} uses an LNES window of 100ms. 
Using one of the other representations with the same window length performs significantly worse.
We also observe that using a shorter temporal window of $33$ms leads to similar performance of the baselines and the LNES window. 
This is expected since less temporal information is lost in shorter baseline windows.
However, our LNES representation supports very long windows ($300$ $ms$) while the accuracy degrades gracefully.
In contrast, the performance of the baselines decays rapidly with increasing window size due to more events being accumulated without any temporal ordering information. 

\begin{figure*}[ht]
    \centering 
    \input{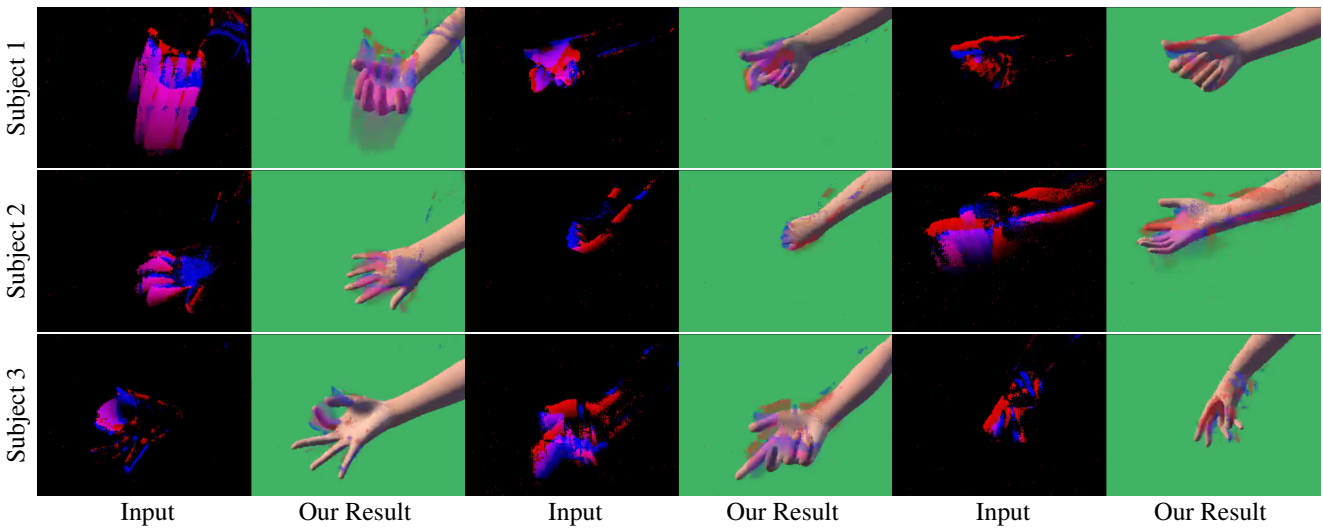}
    \vspace{-0.31cm}
    \caption{
    Results of \emph{EventHands} on real data of different subjects. 
    Our technique predicts a wide variety of hand poses under fast motion.
    }
    \label{fig:resultslong} 
\end{figure*}

\begin{figure*}[ht]
    \centering 
    \input{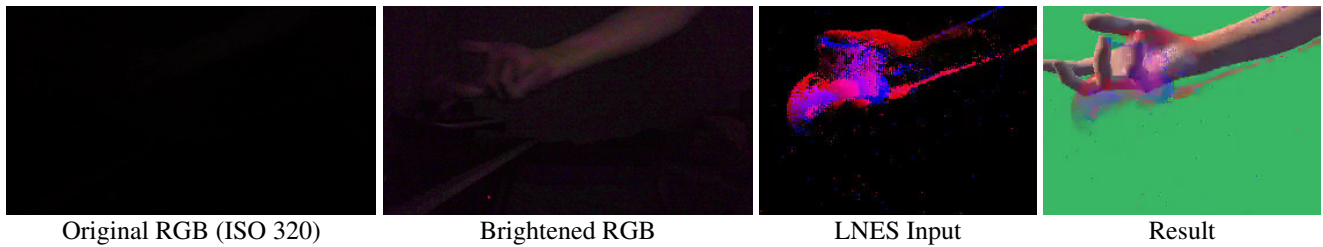}
    \vspace{-0.3cm}
    \caption{
    \emph{EventHands} can reconstruct accurate hand poses in the dark, whereas RGB cameras output starkly under-exposed images. 
    }
    \label{fig:handsinthedark} 
\end{figure*}

\subsection{Comparisons to the State of the Art}
\label{sec:comparisons}

We compare \emph{EventHands} to a variety of RGB techniques~\cite{Zhou2020,GANeratedHands_CVPR2018,Boukhayma_2019_CVPR}. %
Note that data corruptions due to fast motion similarly exist in images produced by commodity depth cameras.
Additionally, we %
show severe failures of a depth-based  state-of-the-art method \cite{Moon2018} in the supplement.

To obtain the input to the RGB techniques at $30$ fps, we apply a moving average filter to the $500$ fps images with a window size of $16$ frames.
Fig.~\ref{fig:compRGB} shows qualitative comparisons against different monocular RGB-based hand pose estimation methods. 
The bounding box estimation of Mueller \textit{et al.}~\cite{GANeratedHands_CVPR2018} is severely impacted by fast motion since they use simple temporal propagation. 
Even if the bounding box 
step succeeds, the blurred input image often leads to erroneous predictions which are propagated to the inverse kinematics skeleton fit.
The method of Boukhayma \textit{et al.}~\cite{Boukhayma_2019_CVPR} also struggles on our test sequences.
In the presence of motion blur, the estimated rigid rotation is inaccurate, and the articulation often defaults to the MANO mean pose (first and last row).
Zhou \textit{et al.}~\cite{Zhou2020} does not predict any hand translation and hence fails in capturing translational movements (first two rows). 
It also fails to capture fast articulations due to motion blur.
Our approach, however, clearly outperforms the RGB-based methods for fast motions. 
Furthermore, \emph{EventHands} also captures fists (third row) where the other methods fail although there is no blur in the input.

For the quantitative comparison, we use the palm-length-normalised 2D-PCKp to ensure a fair comparison across different data modalities.
For Zhou \textit{et al.}~\cite{Zhou2020}, we could not calculate any 2D error since they do not provide translation estimates.
In Fig.~\ref{fig:comparison_real}, we show that our proposed method performs significantly better than the existing RGB-based methods by Mueller \textit{et al.}~\cite{GANeratedHands_CVPR2018} and Boukhayma \textit{et al.}~\cite{Boukhayma_2019_CVPR}.

\vspace{5pt} 

\subsection{Additional Results}
\label{sec:additional_results}
We evaluate \emph{EventHands} on several real videos and show results in Fig.~\ref{fig:resultslong}. 
For each time segment, we show the input event stream and our predicted hand pose overlaid with the input. %
Our approach handles different subjects performing a wide variety of poses and articulations, under fast motion. 

\emph{EventHands} can also handle slow motions without modifications in the network architecture or retraining. 
For that, we detect whether our LNES contains sufficient meaningful (\textit{i.e.}, non-noisy) events and fallback to previous predictions in case of insufficient input events. 
See our supplement for technical details and the video for visualisations.

In contrast to image-based sensors, event cameras monitor relative brightness changes and are able to record reasonable data in dark environments. 
We show one such example 
in Fig.~\ref{fig:handsinthedark} and more in our supplemental video.
\change{We achieve 0.77 2D-PCKp AUC on 236 frames with 1645 annotations. For more details, please refer to the supplementary document.}
\section{Discussion}\label{sec:discussion} 

Our \emph{Event\-Hands} assumes that the scene background is approximately static, \textit{i.e.,} although being robust to a certain degree of noise events, there should not be events in the input that are generated from other moving objects in the scene or due to camera movement.
Although this means that our method is not explicitly designed to handle hand-object and hand-hand interactions, we observe that it is robust to interactions with small objects (see supplementary video).
Future work could investigate how to filter out background events or how to best train a predictor with event data from fully-dynamic scenes.
Another interesting avenue for future research would be to combine both RGB and event data in a way to preserve the low-latency and throughput nature of events, while incorporating the information-rich images where it is easier to detect occlusions and interactions.

Our method would additionally benefit from a learned motion prior to integrate the per-frame predictions better.
Such statistical temporal model could replace the white-noise assumption of the Kalman filter.

\section{Conclusion}\label{sec:conclusions} 
We presented \emph{EventHands}, the first method for 3D hand pose estimation from event streams. 
Our method runs at milestone $1000$ Hz and can reconstruct significantly faster hand motions than any previous work, which is shown in our thorough experiments. 
We believe that the proposed method is also a step forward in general non-rigid 3D reconstruction from event streams, and the presented ideas can be applied in related scenarios and for other types of objects.

\noindent\textbf{Our supplementary material} provides additional results of the proposed approach and further details on the Kalman filter, architecture choice and our event stream simulator. 

\noindent\textbf{Acknowledgements.} This work was supported by the ERC Consolidator Grant 4DRepLy (770784). We thank Jalees Nehvi and Navami Kairanda for help with comparisons.

{\small
\bibliographystyle{ieee_fullname}
\bibliography{egbib}
}

\setcounter{section}{0}
\renewcommand\thesection{\Alph{section}}
\newcommand{\suppsection}{\subsection}
\clearpage
\begin{center}
\textbf{
\Large 
EventHands :: Appendix}
\end{center}
\makeatletter

In this appendix, we provide more insights on our \emph{Event\-Hands} method and additional experimental results. 
First, Sec.~\ref{sec:add_results} discusses details about the experiments and further results. 
Subsequently, we explain the parameters of the Kalman filter in detail   (Sec.~\ref{sec:temporal_filtering}) and the architecture choice for our neural network (Sec.~\ref{sec:net_arch}). 
Sec.~\ref{sec:simulator_details} contains further details about our simulator and the dataset, including the value ranges used for augmentation. 
The supplementary video can be found at \url{https://4dqv.mpi-inf.mpg.de/EventHands/}. 

\section{Experimental Details}
\label{sec:add_results}

\subsection{Baseline Event Representations} 
Here we describe the baseline event representations used for the ablation study. 
 
\noindent\textbf{Event Occurence Image (EOI).}
We define $\mathcal{EOI}  \in \mathbb{R}^{W\times H\times 2} $ to be the event occurence image which is initialised with zeros at the beginning. 
Then, for each event in the current window $\mathcal{E} = \{(t_i, x_i, y_i, p_i)\}_{i=1}^{N_k}$, we update the event occurence image by the following assignment: 
\begin{equation}
    \mathcal{EOI}(x_i, y_i, p_i) = 1. 
\end{equation}
Thus, $\mathcal{EOI}(x_i, y_i, p_i)$ indicates whether an event with polarity $p_i$ has occurred in the window, but it does not consider the temporal event information. 

\noindent\textbf{Single-Channel Event Count Image (ECI-S).}
The single-channel event count image $\mathcal{ECI}\text{-}\mathcal{S} \in \mathbb{R}^{W\times H} $ counts the number of events that occurred at a given pixel, irrespective of their polarity
\begin{equation}
    \mathcal{ECI}\text{-}\mathcal{S}(x, y) = |\{ e_i \in \mathcal{E} \: |\: (x,y) = (x_i, y_i) \}| \: ,
\end{equation}
where $x_i, y_i$ is the position of event $e_i \in \mathcal{E}$.

\noindent\textbf{Event Count Image (ECI).}
Similar to $\mathcal{ECI}\text{-}\mathcal{S}$, the event count image $\mathcal{ECI} \in \mathbb{R}^{W\times H\times 2}$ also counts the number of events in each pixel, however, it contains one channel for each polarity
\begin{equation}
    \mathcal{ECI}(x, y, p) = |\{ e_i \in \mathcal{E} \: |\: (x,y) = (x_i, y_i) \land p = p_i \}| \: ,
\end{equation}
where $x_i, y_i$ is the position of event $e_i \in \mathcal{E}$ and $p_i$ is its polarity.

\subsection{Slow-Motion Settings}
Although the event stream representation is best suited for fast hands, our approach can be adapted, without retraining the model, to also handle slow or stationary hand motions which generate only a small number of events.

Usually, we generate a new LNES with a duration of 100ms every 1ms.
In the slow motion setting, there might not always be enough new events to generate a new LNES. 
When there are fewer than $10$ new events since the last generated LNES, we delay generating a new LNES until at least $10$ new events have happened.
In the case of stationary hands, noise events could eventually accumulate to generate a new LNES frame and cause a random prediction.
We detect this degenerate case by checking the average amount of event information in the last $16$ LNES frames.
The pixel values inside LNES are time stamps and range from 0 (oldest event) to 1 (newest event).
We can hence calculate the total amount of event information in each LNES by summing over all LNES pixel values, which gives more weight to more recent events in each LNES.
If the average amount of event information over the last $16$ LNES is less than $300$, we assume the hand is stationary and repeat the last prediction.
Lastly, we use an additional Kalman filter with the slow setting (see Section 5.3) 
to detect when faster motions occur. If its residual error is ${\geq}0.7$, we switch the main Kalman filter to the fast setting, otherwise we switch it to the slow setting. 
All values are selected empirically.
We show results of this adaptation to slow hands in the supplementary video from 7:30 to 7:45.

\subsection{Additional Results}

\subsubsection{Ablation Studies}
\begin{figure*}
\begin{subfigure}[t]{0.32\textwidth}
    \includegraphics[width=1.0\linewidth]{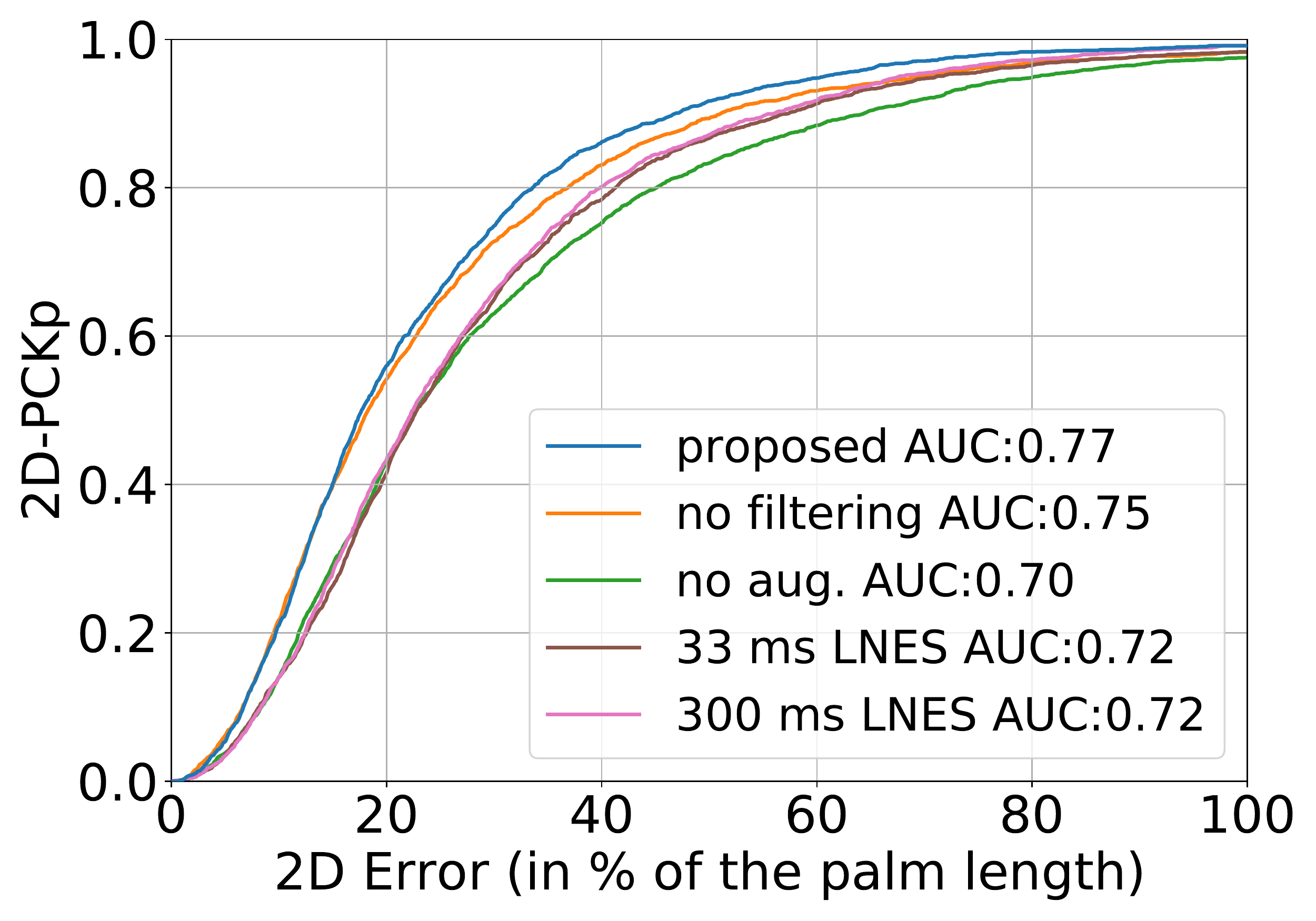}
    \caption{
    Removing filtering leads to a comparably small quantitative decrease in performance whereas removing augmentation has a significant impact on real test data. LNES works well with varying temporal window sizes with the proposed 100ms window achieving the best accuracy.
    }
    \label{fig:ablation_ours}
\end{subfigure} \hfill
\begin{subfigure}[t]{0.32\textwidth}
    \includegraphics[width=1.0\linewidth]{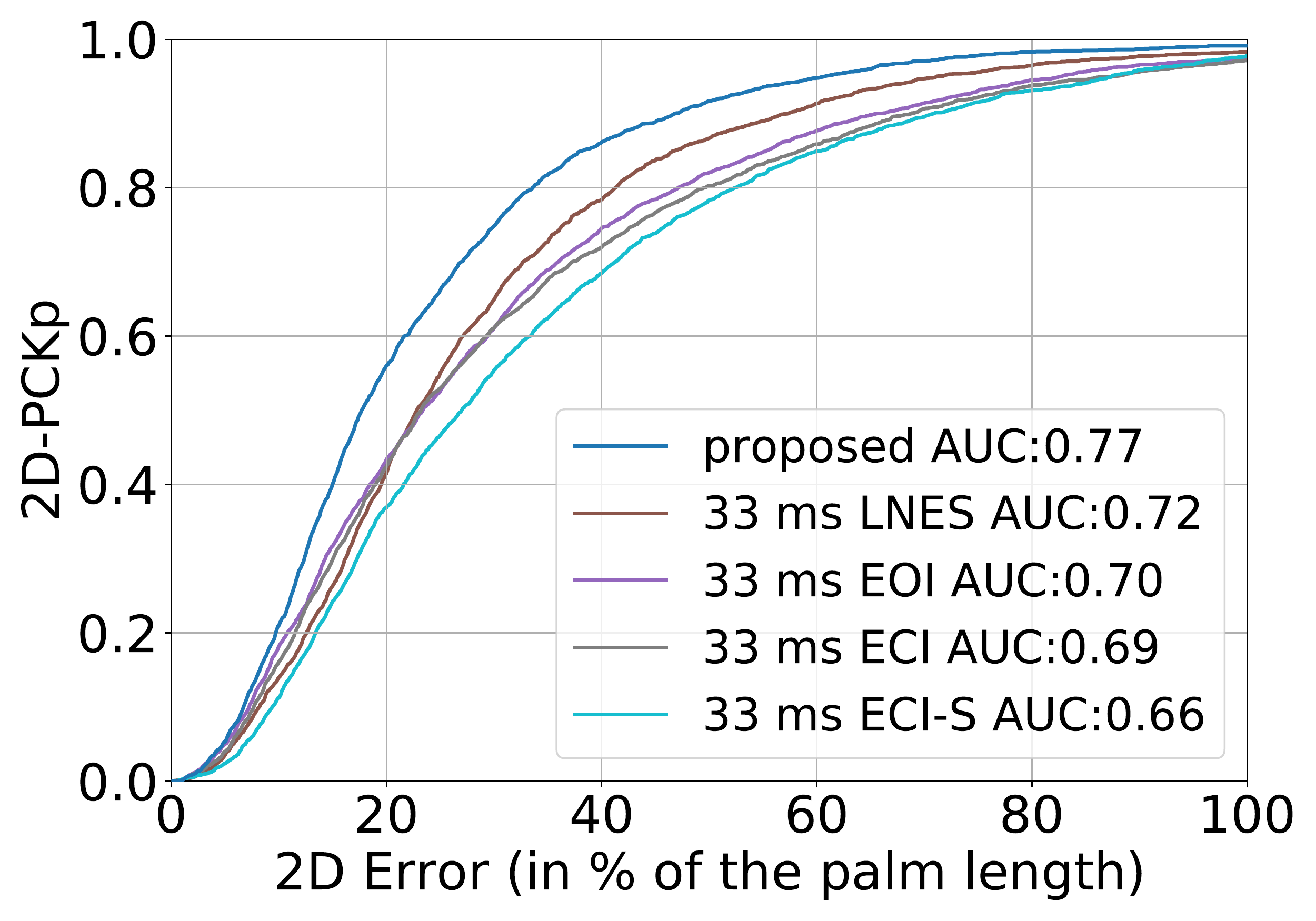}
    \caption{
    With a temporal window size of $33$ms, there is less variation in the performance of the different event representations. Our $33$ms LNES still improves over the other event representations while being less accurate than the proposed $100$ms LNES. 
    }
    \label{fig:ablation_short_window}
\end{subfigure} \hfill
\begin{subfigure}[t]{0.32\textwidth}
    \includegraphics[width=1.0\linewidth]{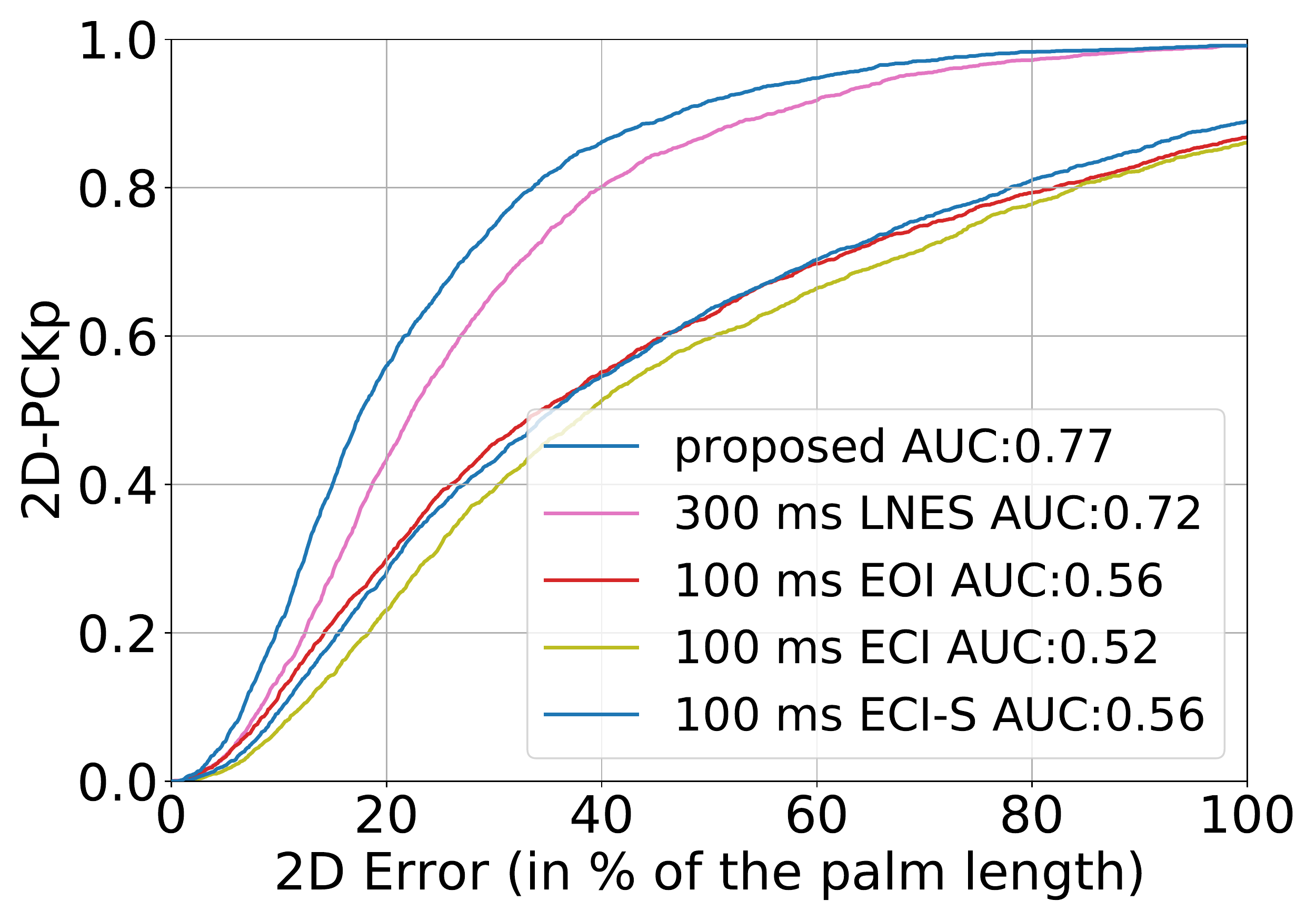}
    \caption{
    When increasing the window size to $100$ms, the difference between LNES and the other representations increases because the latter do not keep any temporal information within the window. While their performance at $100$ms is already significantly degraded, LNES still works with very long windows like $300$ms. 
    }
    \label{fig:ablation_long_window}
\end{subfigure}
    \caption{Quantitative ablation studies on real data. We plot the percentage of keypoints with an error lower than a given threshold. The PCK curves correspond to the AUC values reported in Table 1 in the main paper.}
    \label{fig:ablation_plots}
\end{figure*}

In Fig.~\ref{fig:ablation_plots}, we show the PCK curves corresponding to the AUC values reported in Table 1 
in the main paper.
We compare different settings of our method (no filtering, no augmentation) and different event representations on real test data. The proposed method achieves the best result.
Please refer to Section 6.2 in the main paper for a more detailed discussion.

\subsubsection{Performance of Depth-based Methods}
\begin{figure*}[ht]
    \centering 
    \input{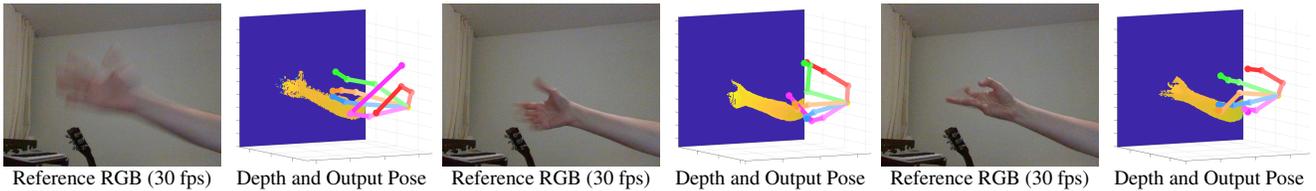}
    \caption{
    Fast moving hands lead to corrupted depth maps on which depth-based methods, like Moon \textit{et al.}~\cite{Moon2018}, produce large errors.
    We show the blurry RGB image for reference only as well as the input depth and the predicted 3D hand pose.
    } 
    \label{fig:depth} 
\end{figure*}

Most commodity depth cameras rely on structured light or time-of-flight techniques to estimate the depth. 
However, for fast motion scenarios targeted by our method, these techniques produce depth estimates that are severely corrupted with many missing depth values. 
As shown in Fig.~\ref{fig:depth}, depth-based state-of-the-art methods such as Moon \textit{et al.}~\cite{Moon2018} cannot handle such artefacts and hence produce erroneous pose estimates. 

\subsubsection{Qualitative Results}

\begin{figure*}
    \centering 
    \input{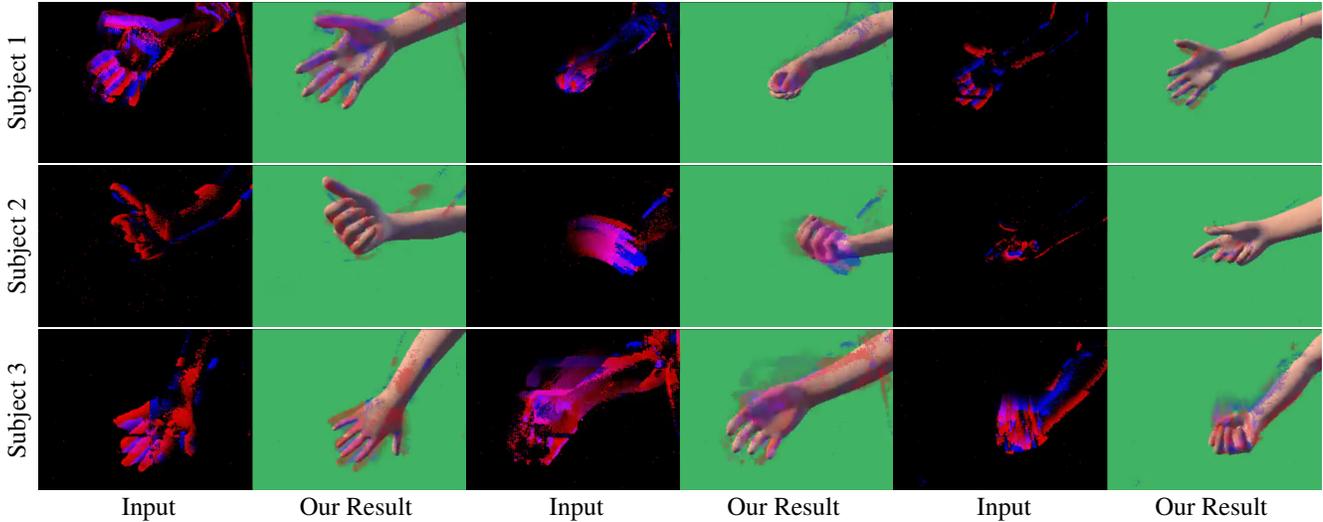}
    \vspace{-0.25cm}
    \caption{
    Additional results of \emph{EventHands} on real event sequences captured with different subjects.
    }
    \label{fig:more_qualitative} 
\end{figure*}

\begin{figure*}
    \centering 
    \input{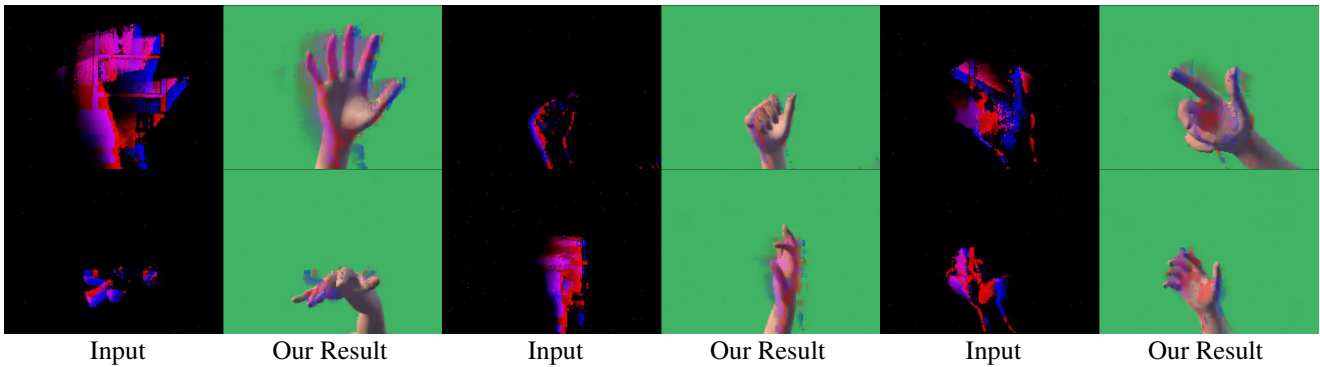}
    \vspace{-0.25cm}
    \caption{
    Results of \emph{EventHands} on real data where the hand is entering the frame from the bottom.
    }
    \label{fig:qualitative_botom} 
\end{figure*}

Fig.~\ref{fig:more_qualitative} shows more qualitative results for different subjects that we captured with the DAVIS240C event camera (\emph{EventHands} uses event stream only). 
Furthermore, we provide results of a network trained with the arm entering the  field of view from the bottom in Fig.~\ref{fig:qualitative_botom}. 
In this experiment, we use additional $55$ hours of generated event stream data for training. 

\subsubsection{Low-Light Performance}
\change{We also annotated a 7 second part of the recorded low-light event stream. The annotations were done the same way as described in Section~6.1, with 236 frames and 1645 keypoints annotated in total. For visual results of our method on the extended material, please refer to the supplementary video (at 07:45).}

\section{Temporal Filtering}
\label{sec:temporal_filtering}

We use a Kalman filter \cite{labbe2014kalman} with constant velocity assumption to post-process our raw predictions $\theta \in \mathbb{R}^{12}$. 
The corresponding state vector $S \in \mathbb{R}^{24}$ is given by 
\begin{equation} 
S =  
\begin{bmatrix}
    \theta_{1}&
    \dot{\theta}_{1}&
    \hdots&
    \theta_{12}&
    \dot{\theta}_{12}
\end{bmatrix}^T,
\end{equation} 
where $\dot{\theta}_i$ is the velocity of $i$-th parameter $\theta_i$. 
We model changes in velocities $\dot{\theta}_i$ as independent Gaussian white noise (\textit{i.e.,} temporally uncorrelated). 
For a given process noise variance $\sigma_P^2$, the  \textit{discrete  white noise covariance matrix operator}  produces a block-diagonal covariance matrix 
\begin{equation}\label{eq:white_noise_operator} 
\omega(\sigma_P^2)= \sigma_P^2 
\begin{pmatrix}
    \diagentry{W_{1}}\\
    &\diagentry{\xddots}\\
    &&\diagentry{W_{12}}
\end{pmatrix}. 
\end{equation} 
This matrix models uncertainty in updating both the position $\theta$ and velocity $\dot{\theta}$ in the state vector $S$. 
In Eq.~\eqref{eq:white_noise_operator}, $W_{i}$ is the process noise  covariance matrix of $[\theta_i, \dot{\theta_i}]$: 
\begin{equation} 
W_i = 
\begin{pmatrix}
    \frac{1}{4}\Delta t^4 & \frac{1}{2}\Delta t^3\\[0.3em]
    \frac{1}{2}\Delta t^3 & \Delta t^2\\[0.3em]
\end{pmatrix}, 
\end{equation} 
where $\Delta t$ is the temporal step size. 

\section{Choosing Network Architecture} 
\label{sec:net_arch}
Besides ResNet-18, we examined other base models including  \mbox{VGG-\{11,13,16,19\}} (with batch normalisation)~\cite{vgg}, 
\mbox{MobileNet v2}~\cite{mobilenet},
\mbox{ShuffleNet v2}~\cite{shufflenet}, 
\mbox{Inception v3}~\cite{inception}, \mbox{MnasNet}~\cite{mnasnet} and \mbox{ResNet-34}. 
Out of these models, only \mbox{MobileNet v2}, \mbox{ResNet-34} and VGG networks produced validation losses comparable to \mbox{ResNet-18}.
However, the smallest examined VGG network and \mbox{ResNet-34} were unable to handle real-time processing at 1000~Hz, which was one of our main goals. 
Furthermore, while the inference time of \mbox{MobileNet v2} was faster than that of ResNet-18, we select ResNet-18 as the base model as it has higher prediction accuracy and enables $1000$ frames per second. 

\section{Simulator and Dataset Details}
\label{sec:simulator_details}
This section provides more details on the implementation of our GPU-based event simulator and the format used for storing our dataset. 

\subsection{Implementation Details}
The simulator is developed in C++ using
\begin{itemize}
    \item CUDA, cuBLAS, cuRAND --- for computing MANO pose-corrective mesh offsets (that reduce skinning artefacts) on GPU and event camera simulation, 
    \item OpenGL --- for posing the skinned and corrected MANO mesh as well as rendering the scene, 
    \item xtensor --- for computing MANO shape template mesh and textures on CPU and for loading MANO data, and 
    \item SDL2 --- for image loading and OpenGL context management.
\end{itemize}

All mesh operations that happen every frame, are performed entirely on GPU using GPU memory only. 
Only two CPU-GPU memory transfers are needed per frame to obtain the current pose vector and the event stream output. 

This and other optimisations allow fast simulation of the full SMPL+H body model at $240{\times}180$ resolution, which is the resolution of the DAVIS240C event camera that we use for the experiments. 
On a single NVIDIA GeForce GTX1070, two instances of the simulator can be launched simultaneously to obtain events at rates of around $2000$ simulated time steps per second. Considering that we use a time step equal to $1/1000$ of a second, that means we can simulate data at twice the real-time speed with $1$ $ms$ temporal resolution.

\subsection{Event Camera Calibration} 
To reduce the domain gap between the simulated and real event data, we used the event threshold value and the noise event rate of our DAVIS240C event camera.

To calibrate the event threshold $C$, we shot several sequences by moving an object (a checkerboard) monotonously from one side to the other with different speeds. 
We captured both the events $\{(t_i, u_i, p_i)\}_{i=1}^{N_\mathrm{events}}$ and instant intensity images $\{\Omega_j\}_{j=1}^{N_\mathrm{images}}$ simultaneously. 
By moving monotonously from one side to the other side, we eliminate cases when the event stream contains events that cannot be explained by the intensity images, \textit{e.g.,} events that cancel themselves between two consecutive intensity images. 
To estimate the event threshold from the captured data, we use the following observation. 
According to our camera model, if the camera emits $N$ events in total, 
then the intensity images would have the total log-intensity change of ${\approx}NC$. 
Thus, $C$ can be estimated by dividing the total log-intensity change by the total number of events $N$. 

Hence, we counted the total intensity change of the instant intensity images $\Delta_\mathrm{total}$ as 
\begin{equation}
\small
\Delta_\mathrm{total} = \sum_{i=1}^{N_\mathrm{images}-1} {| \log(\max\{\Omega_{t+1},\varepsilon\})-\log(\max\{\Omega_t,\varepsilon\}) |},
\end{equation}
where $\varepsilon=10$ is a constant added for numerical stability. 
Then, we estimate $C$ as
\begin{equation}
    C \approx \Delta_\mathrm{total}/N_\mathrm{events}. 
\end{equation}
For our event camera, we obtain 
$C = 0.5$--$0.55$. 

To estimate the \emph{noise event rate}, %
we shot the static background and count the number of positive and negative recorded events. 
For our DAVIS240C, we estimate the noise to be ${\approx}2500$ positive and ${\approx}100$ negative events per second. 

\subsection{Dataset Format}
The generated dataset consists of two files, \textit{i.e.,} the event stream and the metadata stream. 
The event stream file format is tailored for the frame-by-frame event stream simulation. 
It consists of blocks of four bytes: 
two bytes for $x$ coordinate, one byte for $y$ coordinate and one byte for polarity $p$. 
At the start, the timestamp is considered to be zero. 
A new frame is indicated by the polarity value $p=255$, which signals that the timestamp should be incremented by one time step. 
We use the time step of $1/1000$ of a second. 
The file starts with a four-byte integer that specifies the number $N$ of metadata fields per each frame. 
Then, the stream starts and it consists of $8N+2$ byte blocks. 
The block contains $N$ eight-byte double-precision reals and two-byte magic. 
We use $N = 12$ for six MANO articulation coefficients,  three  components of the hand root translation vector,  and three components of the hand root rotation vector. 

We also implement a high-speed C++/Python loader for the proposed dataset format. 
It allows loading ${\sim}8.6 \cdot 10^5$ simulated frames per second or ${\sim}1.75 \cdot 10^8$ events per second when using storage capable of $1000$ MB/s read speeds. With the fixed rate of $1000$ simulated frames per second, this amounts to loading $860$ simulated seconds per second. 
Thus, we are able to load a $45$-hours-long dataset in just three minutes.

\subsection{Simulation Parameters} 
We next describe how we augment the simulation for generating the event data. 
SMPL+H body shape $\beta$ is drawn from $\mathcal{U}[-2, 2]$. 
Body position $\theta$ is drawn as follows. First, we sample $\xi \sim \mathcal{U}[-0.2, 0.2]$. Then $\theta=\xi\odot g+o$, where $g$ is the gain vector, $o$ is the offset vector and $\odot$ is the component-wise multiplication operator. 

For the dataset in which the arm comes from the top and right edges, the gain is
\[
g^{(1)}_i=\begin{cases}
100, &\textrm{if }i=16\cdot 3+9,\\
40, &\textrm{if }i=16\cdot 3+10,\\
10, &\textrm{if }i=16\cdot 3+11,\\
40, &\textrm{if }i=16\cdot 3+15,\\
40, &\textrm{if }i=16\cdot 3+16,\\
40, &\textrm{if }i=16\cdot 3+17,\\
1, &\textrm{otherwise,}
\end{cases}
\]
and the offset is
\[
o^{(1)}_j=\begin{cases}
0.2, &\textrm{if }j=13\cdot 3+5,\\
0.1, &\textrm{if }j=16\cdot 3+5,\\
1.4 \epsilon, &\textrm{if }j=16\cdot 3+9,\\
0.5, &\textrm{if }j=16\cdot 3+11,\\
0, &\textrm{otherwise,}
\end{cases}
\]
where $\epsilon$ is sampled randomly and is either $-1$ or $1$ with equal probability.
Global translation vector has $(x, y)$ components sampled from $\mathcal{U}[-0.3, 0.3]$. The depth component $z$ is taken from $\mathcal{U}[-0.09, 0.09]$. 

For the dataset in which the arm comes from the bottom edge, the gain is
\[
g^{(2)}_i=\begin{cases}
100, &\textrm{if }i=16\cdot 3+9,\\
10, &\textrm{if }i=16\cdot 3+11,\\
40, &\textrm{if }i=16\cdot 3+15,\\
40, &\textrm{if }i=16\cdot 3+16,\\
40, &\textrm{if }i=16\cdot 3+17,\\
1, &\textrm{otherwise,}
\end{cases}
\]
and the offset is
\[
o^{(2)}_j=\begin{cases}
-2.3, &\textrm{if }j=2,\\
0.2, &\textrm{if }j=13\cdot 3+5,\\
0.1, &\textrm{if }j=16\cdot 3+5,\\
1.4 \epsilon-3.8, &\textrm{if }j=16\cdot 3+9,\\
0.5, &\textrm{if }j=16\cdot 3+11,\\
0, &\textrm{otherwise,}
\end{cases}
\]
where $\epsilon$ is also chosen randomly and is either $-1$ or $1$ with equal probability. 
Global translation vector has $x$ component drawn from $\mathcal{U}[-1.5, -0.9]$, $y$ component taken from $\mathcal{U}[-0.52, 0.08]$ and depth component $z$ taken from $\mathcal{U}[-0.09, 0.09]$. 

The hand MANO articulation parameters are sampled from $\mathcal{U}[-2, 2]$, whereas hand texture PCA coefficients are chosen from $\mathcal{N}(0, 4I)$. 
Light directions are sampled uniformly from all possible directions and light intensities are drawn from $\mathcal{U}[0.9,1.1]$. 
Finally, the background image is drawn randomly from the  collected set of nine background images. 
The event generation threshold is drawn from $\mathcal{N}(0.5, 0.0004)$.

\end{document}